\documentclass{article} 
\usepackage{iclr2027_conference,times}

\usepackage{amsmath,amsfonts,bm}

\def\eqref#1{equation~\ref{#1}}

\def\1{\bm{1}}

\DeclareMathAlphabet{\mathsfit}{\encodingdefault}{\sfdefault}{m}{sl}
\SetMathAlphabet{\mathsfit}{bold}{\encodingdefault}{\sfdefault}{bx}{n}

\usepackage{hyperref}
\usepackage{url}
\hypersetup{
  colorlinks=true,
  allcolors=blue
}
\usepackage{graphicx}
\usepackage{booktabs}
\usepackage{wrapfig}
\usepackage{xcolor}
\usepackage{pifont}
\usepackage{multirow}
\usepackage{enumitem}
\usepackage{placeins}

\newcommand{\fitwidth}[2]{%
  \sbox0{#2}%
  \ifdim\wd0>#1 \resizebox{#1}{!}{\usebox0}%
  \else \usebox0\fi}

\title{World2Motion: Turning Video World Models into 3D Human Motion Generators}

\author{\bfseries Fangyuan Tu$^{1}$, Xiangyue Zhang$^{2}$, YIYI CAI$^{2}$, YICHEN PENG$^{4}$ \\
\bfseries Kunhang Li$^{2}$, Bo Zheng$^{3}$, Zhixiang Wang$^{3}$, Kaipeng Zhang$^{3}$ \\
\bfseries Erwin Wu$^{4}$, Haoran Xie$^{1}$, Haiyang Liu$^{2}$ \\[4pt]
{\normalfont\small $^{1}$Japan Advanced Institute of Science and Technology} \\
{\normalfont\small $^{2}$The University of Tokyo \quad $^{3}$Alaya Lab} \\
{\normalfont\small $^{4}$Institute of Science Tokyo}
}

\iclrfinalcopy
\begin{document}

\maketitle
\fancyhead{}
\lhead{Preprint}

\begin{abstract}
\begin{figure}[h]
    \centering
    \includegraphics[width=\linewidth]{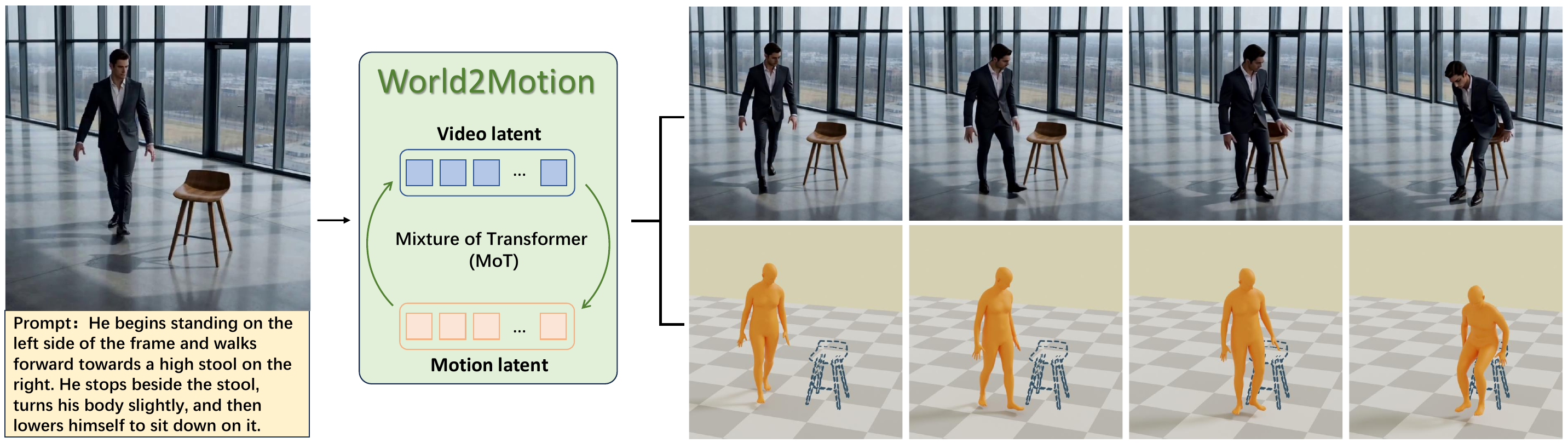}
    \caption{World2Motion adapts a video world model to jointly generate scene-aware 3D human motion and corresponding video from a single image and a text prompt. The example shows a person approaching a stool, turning, and sitting down. The top row shows generated video frames, while the bottom row shows the corresponding 3D motion rendered in a scene for visualization.}
    \label{fig:teaser}
\end{figure}

We present World2Motion, a framework that generates scene-aware 3D human motion and corresponding video from a single image and a text prompt. While existing 3D motion generators learn from motion datasets, their generalization is constrained by limited coverage of environments. In contrast, video world models such as Cosmos 3 offer broader environmental priors but are not designed for full-body motion generation; recovering motion from their generated videos requires costly two-stage inference. To address these, we turn Cosmos 3 into a single-stage 3D motion generator. This adaptation has two challenges: the scarcity of paired video--motion data and temporal instability in the generated motion. First, we construct a training dataset combining synthetic video--motion pairs with real videos paired with estimated 3D motion. Second, we propose a shift-decoupled noise schedule that assigns different noise levels to video and motion through shared denoising progress. This design accommodates the different denoising requirements of the two modalities, reducing motion jitter. Experiments on a multi-source interaction benchmark show that World2Motion has better motion--text alignment and scene interaction compared with the evaluated 3D motion generators. It also matches the interaction success rate of the two-stage baseline while achieving approximately 3.3$\times$ faster inference. Our project page is available at \url{https://fyantu.github.io/World2Motion/}. 


\end{abstract}

\section{Introduction}
\label{sec:intro}
Generating 3D human motion for interactive applications requires motions that follow text descriptions and remain consistent with the surrounding environment. Conventional text-to-motion (T2M) methods~\citep{zhang2024motiondiffuse,tevet2023mdm} synthesize human motion from textual descriptions, and recent approaches~\citep{hymotion2025,Kimodo2026} have improved generation quality by scaling up motion datasets. However, the limited coverage of environments and interactions in available 3D motion datasets constrains their generalization to diverse scenes. Text alone also makes it difficult to specify detailed spatial relationships. For example, generating a person sitting on a chair requires the motion to match the chair's position and orientation. Human--object interaction (HOI) methods~\citep{li2024chois,jung2026dechoi} incorporate such spatial information through explicit 3D inputs, such as object meshes~\citep{kulkarni2024nifty} or object states~\citep{li2024chois,zhang2026giraf}. However, collecting the corresponding motion and scene data at scale remains difficult.

Video world models offer an alternative source of knowledge about human motion and environmental interactions~\citep{wan2025wan,minimaxh3}. Trained on diverse videos, models such as Cosmos 3~\citep{agarwal2026cosmos} provide environmental priors that can complement the limited coverage of 3D motion datasets. However, they are not designed to directly produce the full-body 3D motion required by our task. A straightforward way to use these priors is to first generate a video and then recover motion using a human mesh recovery (HMR) model~\citep{patel2025camerahmr,zhao2026onlinehmr}. This two-stage pipeline enables scene-aware motion generation, but requires a separate motion-recovery pass over the generated video, increasing inference cost.



\begin{wraptable}{r}{0.50\textwidth}
\centering
\normalfont\small\color{black}
\setlength{\tabcolsep}{2pt}
\caption{Motion generation methods comparison.}
\label{tab:paradigm comparison}
\begin{tabular*}{\linewidth}{@{\extracolsep{\fill}}lccc@{}}
\toprule
Approach & \shortstack{Direct 3D\\motion\\output} & \shortstack{No 3D\\object\\input} & \shortstack{Scene\\conditioning} \\
\midrule
T2M & \ding{51} & \ding{51} & \\
I2V + HMR & & \ding{51} & \ding{51} \\
HOI & \ding{51} & & \ding{51} \\
World2Motion & \ding{51} & \ding{51} & \ding{51} \\
\bottomrule
\end{tabular*}
\end{wraptable}

We introduce World2Motion, which adapts the existing Mixture-of-Transformers (MoT) backbone of Cosmos 3~\citep{agarwal2026cosmos} to jointly generate scene-aware 3D human motion and corresponding video from a single image and a text prompt. This single-stage framework avoids a separate motion-recovery stage on the generated sequence. The adaptation presents two challenges: the scarcity of paired video--motion training data and temporal instability in the generated motion. Table~\ref{tab:paradigm comparison} summarizes the input requirements and generation pipelines of these approaches.

To address the scarcity of paired supervision, we construct a hybrid training dataset combining synthetic video--motion pairs with real videos paired with estimated 3D motion. For the synthetic component, we use motion sequences from existing interaction datasets~\citep{li2023omomo,hassan2021samp,jiang2023chairs} to construct corresponding videos, providing paired supervision with known 3D motion. For the real component, we estimate 3D motion from real videos~\citep{patel2025camerahmr} to expand the coverage of environments and appearances. These two sources combine motion supervision from existing 3D datasets with the visual diversity of real videos, supporting the adaptation of the video world model to joint video--motion generation.

Directly fine-tuning the backbone for joint video--motion generation produces noticeable motion jitter. To improve temporal stability, we employ root translations relative to the initial root position and 6D joint rotations~\citep{zhao2026ardy}. We propose a shift-decoupled noise schedule that assigns different noise levels to video and motion through shared denoising progress, accounting for their different denoising requirements. We maintain the same video--motion noise relationship during training and inference. Our ablation studies show that the shift-decoupled noise schedule reduces motion jitter.

On a multi-source interaction benchmark, World2Motion improves motion--text alignment and scene interaction over the evaluated 3D motion generators. It also achieves comparable interaction success to the MiniMax-H3 + CameraHMR baseline with approximately 3.3$\times$ faster inference. 

Our contributions are as follows:
\begin{itemize}[leftmargin=*]
\item We adapt Cosmos 3 for single-stage joint generation of scene-aware 3D human motion and corresponding video from a single image and a text prompt, avoiding a separate motion-recovery stage on the generated video.
\item We construct a hybrid training dataset combining synthetic video--motion pairs with real videos paired with estimated 3D motion.
\item We propose a shift-decoupled noise schedule that uses modality-specific noise shifts with shared denoising progress and maintains the same video--motion noise relationship during training and inference, improving temporal stability.
\end{itemize}

\section{Related Work}
\label{sec:related}

\noindent \textbf{Text-conditioned human motion generation}. Text-to-motion (T2M) methods generate 3D human motion from natural language. Diffusion-based methods such as MotionDiffuse~\citep{zhang2024motiondiffuse} and MDM~\citep{tevet2023mdm} denoise motion sequences, while MLD~\citep{chen2023mld} performs diffusion in a learned motion latent space and ReMoDiffuse~\citep{zhang2023remodiffuse} incorporates retrieved motion examples. Autoregressive methods, including T2M-GPT~\citep{zhang2023t2mgpt} and MotionGPT~\citep{jiang2023motiongpt}, generate discrete motion tokens sequentially. Masked models such as MoMask~\citep{guo2024momask}, BAMM~\citep{pinyoanuntapong2024bamm}, and MaskControl~\citep{pinyoanuntapong2025maskcontrol} further explore token prediction for motion generation and control. MoMask uses residual vector quantization to represent motion with multiple token layers, while MaskControl incorporates spatial constraints. Recent work has also studied larger training datasets and more flexible motion constraints. HY-Motion~\citep{hymotion2025} scales DiT-based flow matching to a billion parameters, while Kimodo~\citep{Kimodo2026} trains a motion diffusion model on large-scale optical motion capture data and supports pose and trajectory constraints. These methods improve motion generation and control, but do not directly use scene images to model spatial relationships between a person and the surrounding environment.

\noindent \textbf{Scene-conditioned human motion generation.} Scene-conditioned methods incorporate environmental information into human pose and motion generation. Methods such as POSA~\citep{hassan2021posa}, SAMP~\citep{hassan2021samp}, COINS~\citep{zhao2022coins}, and HUMANISE~\citep{wang2022humanise} use 3D scene information to model human placement, contact, or motion. Other approaches explore different forms of interaction conditions: OMOMO~\citep{li2023omomo} uses object trajectories, Move as You Say~\citep{wang2024move} uses scene affordances, and TRUMANS~\citep{jiang2024trumans} conditions motion generation on scene context and actions. SceneDiffuser~\citep{huang2023scenediffuser} and InfBaGel~\citep{zou2026infbagel} further study scene-conditioned generation and refinement for human interactions. These methods explore how scene information can guide motion beyond textual descriptions. Move-in-2D~\citep{huang2025movein2d} generates motion from a background image and text, and uses the generated motion to guide subsequent video generation. Our work studies the adaptation of a pretrained video world model for image-conditioned joint video and 3D motion generation.

\noindent \textbf{Human motion recovery from images and videos.} Human motion recovery estimates 3D body pose and shape from observed images or videos, often using a parametric body model such as SMPL~\citep{SMPL:2015}. HMR~\citep{kanazawa2018hmr}, SPIN~\citep{kolotouros2019spin}, CLIFF~\citep{li2022cliff}, and HybrIK~\citep{li2021hybrik} improve recovery through learned pose priors, optimization during training, image-location cues, and kinematic constraints. More recent methods study transformer-based recovery, whole-body and multi-person reconstruction, and camera and temporal modeling, including Humans in 4D~\citep{goel2023humansin4d}, SMPLer-X~\citep{cai2023smpler}, Multi-HMR~\citep{baradel2024multihmr}, CameraHMR~\citep{patel2025camerahmr}, MetricHMSR~\citep{song2026metrichmsr}, and OnlineHMR~\citep{zhao2026onlinehmr}. In our framework, CameraHMR provides motion estimates for real training videos and the initial pose from the input image. Motion recovery can also be applied to generated videos, forming a two-stage motion-generation pipeline, but the additional recovery stage increases inference cost. In contrast, World2Motion jointly generates motion and video, achieving approximately $3.3\times$ faster inference, better motion--text alignment, and lower motion jitter than the MiniMax-H3~\citep{minimaxh3} + CameraHMR~\citep{patel2025camerahmr} baseline.

\noindent \textbf{Video world models and joint video--motion generation.} Large-scale video models, such as Wan~\citep{wan2025wan} and MiniMax-H3~\citep{minimaxh3}, learn visual priors about human motion and environmental interactions. Cosmos 3~\citep{agarwal2026cosmos} extends multimodal generation to video and action within a Mixture-of-Transformers architecture. Its action modeling includes egocentric head and hand motion, and its action training inherits the vision noise schedule. This provides a starting point for adapting video world models to full-body 3D motion generation. Related work has also explored joint video--motion generation. CoMoVi~\citep{zhao2026comovi} uses separate diffusion branches for RGB video and 2D motion representations, together with 3D--2D cross-attention to predict 3D motion within the same denoising loop. World2Motion adapts the existing multimodal backbone of Cosmos 3 for joint video--motion generation. We combine synthetic and real video--motion training pairs and propose a shift-decoupled noise schedule that assigns modality-specific noise levels through shared denoising progress to improve temporal stability.

\section{Method}
\label{sec:method}

\begin{figure}[!htbp]
    \centering
    \includegraphics[width=\linewidth]{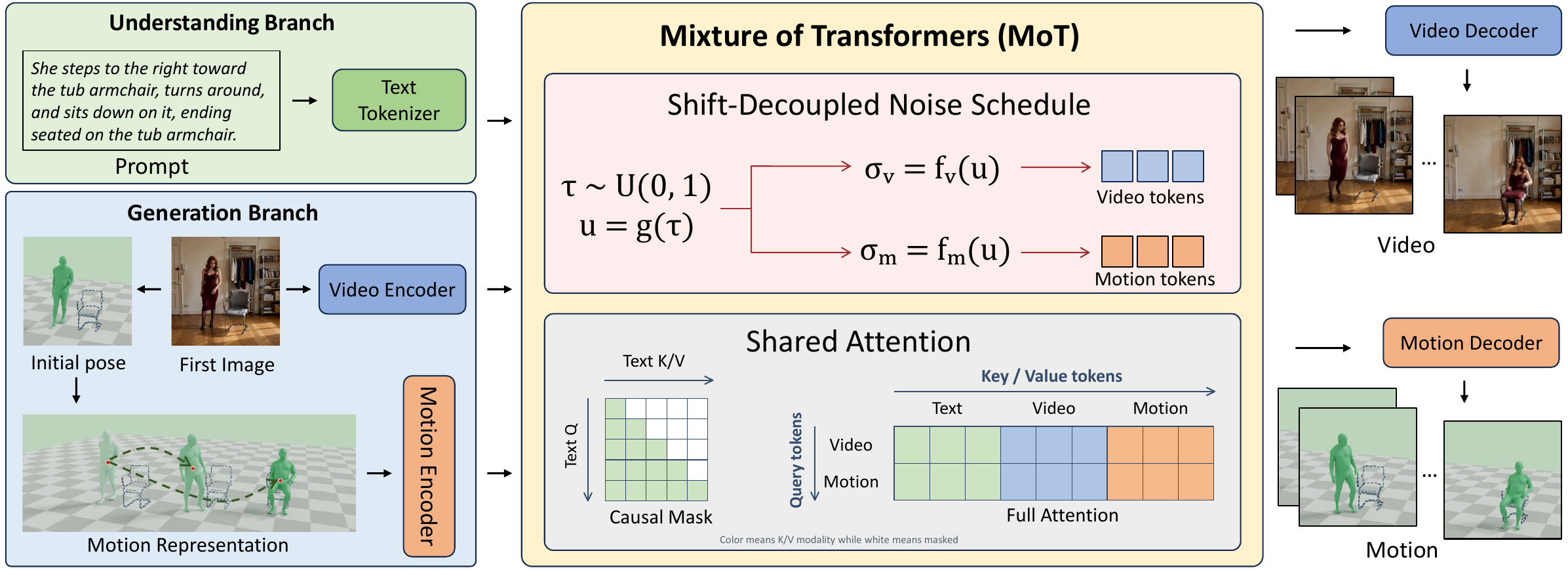}
    \caption{Overview of World2Motion. From an image and a text prompt, the adapted Cosmos 3 backbone jointly generates 3D human motion and video. CameraHMR estimates the initial pose from the image. The shift-decoupled schedule uses modality-specific noise shifts with shared denoising progress.}
    \label{fig:Method}
\end{figure}

Given a single image containing a person and a text prompt describing the intended action, World2Motion jointly generates a video sequence and corresponding 3D human motion. We estimate the initial human pose from the input image using CameraHMR~\citep{patel2025camerahmr}. To use the priors learned during video world model pretraining, we adapt the existing Cosmos 3 backbone for full-body motion generation. Figure~\ref{fig:Method} provides an overview of the framework.

\subsection{Review of Cosmos 3}
Cosmos 3~\citep{agarwal2026cosmos} uses a Mixture-of-Transformers (MoT) backbone with an understanding branch and a generation branch. We retain its text and video encoding components, shared attention mechanism, positional encoding, and train both generated modalities with rectified-flow objectives.

\subsection{Hybrid Video--Motion Training Data}
\label{sec:training-data}
To address the scarcity of paired supervision, we construct HOI-mix, a hybrid dataset combining synthetic video--motion pairs with real videos paired with estimated 3D motion. For the synthetic component, we render motion sequences from existing interaction datasets, including OMOMO~\citep{li2023omomo}, CHAIRs~\citep{jiang2023chairs}, and SAMP~\citep{hassan2021samp}, and use the rendered animations to guide MiniMax-H3~\citep{minimaxh3} in generating corresponding RGB videos. The original motion sequences provide known 3D supervision. For the real component, we estimate 3D motion from curated Bilibili videos using CameraHMR. We use Gemini 2.5~\citep{comanici2025gemini} to generate text descriptions for both components. These sources combine motion supervision from existing 3D datasets with the diversity of environments and appearances in real videos.

\subsection{Full-Body Motion Adaptation}
We represent each SMPL~\citep{SMPL:2015} motion frame using root translation and 6D joint rotations, including the root orientation. This representation allows generation to be conditioned on an initial pose without requiring velocity estimates. Following ARDY~\citep{zhao2026ardy}, we subtract the initial root position from the root translation at each frame. We convert the axis-angle joint rotations to 6D rotations, normalize the combined translation and rotation features, and project them into the Transformer hidden dimension using a motion encoder. A learnable modality embedding distinguishes motion tokens from video tokens. Motion time coordinates are scaled to match the video VAE's fourfold temporal compression, aligning the two modalities in time.

We train the motion encoder and decoder from scratch and initialize the remaining model weights from Cosmos 3. The understanding branch remains frozen during training. Our ablation study compares relative and absolute root translations and shows lower jerk and foot sliding with the relative representation (Table~\ref{tab:ablation_norm}).

\begin{figure}[!tb]
    \centering
    \includegraphics[width=\linewidth]{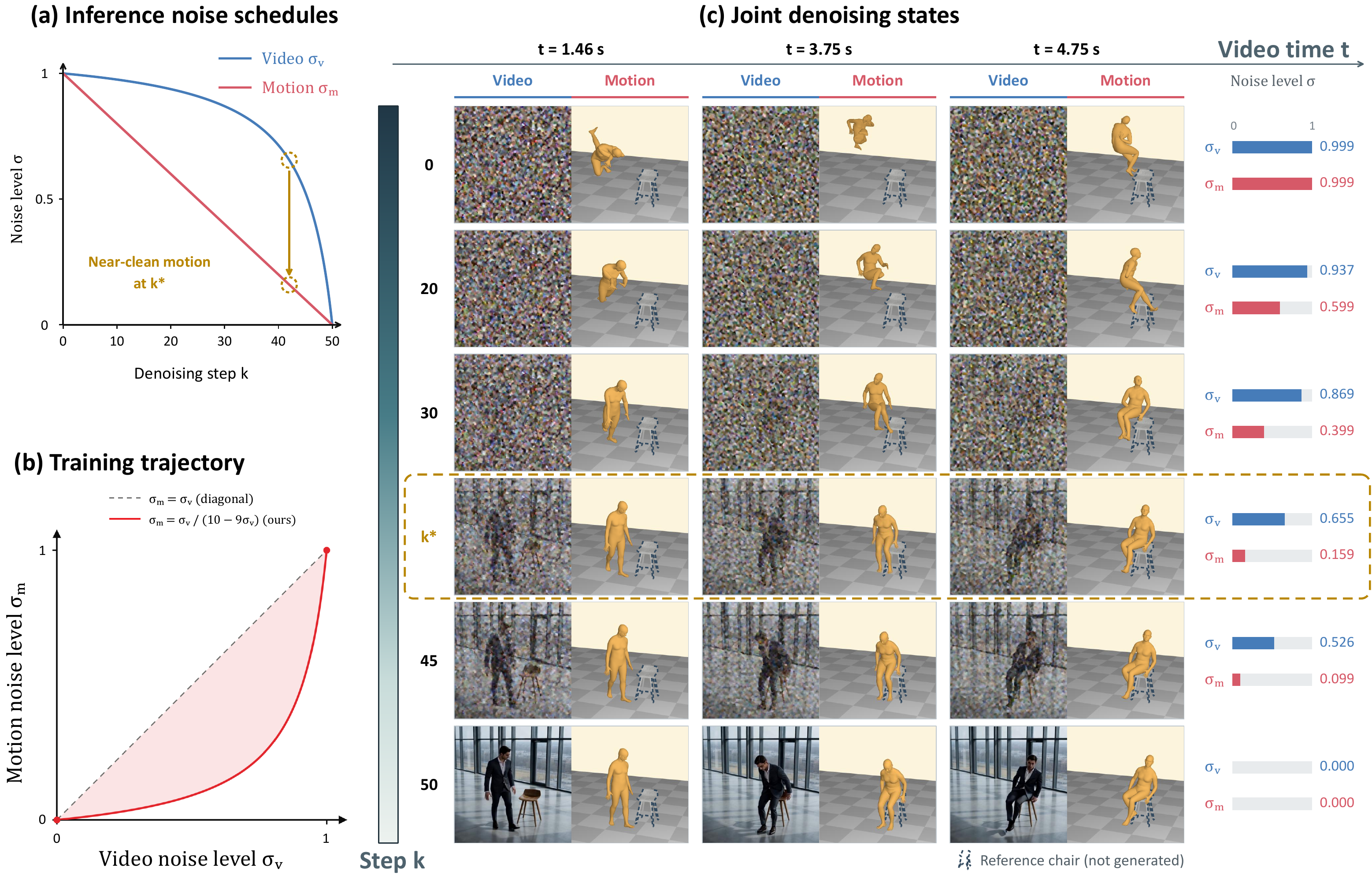}
    \caption{Shift-decoupled video--motion denoising. (a) Video and motion noise levels during inference with $s_v=10$ and $s_m=1$. (b) The corresponding training noise pairs follow a shared curve, with lower motion noise than video noise at intermediate steps. (c) Intermediate video frames and motion states. Rows show denoising steps, and column groups show video times. The chair mesh is reference geometry for visualization.}
    \label{fig:denoising}
\end{figure}

\subsection{Shift-Decoupled Noise Schedule}
To improve temporal stability in joint video--motion generation, we propose a shift-decoupled noise schedule. It uses separate noise shifts for video and motion while retaining shared denoising progress $u\in[0,1]$. At each training step, we sample one value of $u$ and compute the video and motion noise levels as
\begin{equation}
\sigma_v(u) = \frac{s_vu}{1+(s_v-1)u}, \qquad
\sigma_m(u) = \frac{s_mu}{1+(s_m-1)u},
\label{eq:coupled-noise-schedule}
\end{equation}
where $s_v$ and $s_m$ are positive shifts for video and motion, respectively. The shifts control how shared progress maps to each modality's noise level. Setting $s_v>s_m$ keeps video at a higher noise level than motion during intermediate denoising steps, as illustrated in Figure~\ref{fig:denoising}(a). Each modality receives a timestep embedding corresponding to its own noise level. The shared progress constrains the training noise pairs to a single curve in the video--motion noise space, as illustrated in Figure~\ref{fig:denoising}(b).

During inference, both modalities follow a common progress grid from $u=1$ to $u=0$, using the same shifts and mappings as in training. Both noise schedules reach zero at the final step. Among the tested configurations, $s_v=10$ and $s_m=1$ achieve the lowest jerk and foot sliding rate. Matching the inference shifts to the training shifts also improves both metrics compared with mismatched settings (Table~\ref{tab:ablation_schedule}).

After joint generation, we inverse-normalize the motion output and recover global root translation by adding the initial root position to the predicted relative translation. We then apply a Savitzky--Golay filter~\citep{savitzky1964smoothing} as post-processing. We disable this filter in all ablation experiments to evaluate the effects of the motion representation and noise schedule directly.

\section{Experiments}
\label{sec:experiments}
\subsection{Experimental Setup}
\label{sec:experimental-setup}

\noindent\textbf{Dataset Construction and Splits.} Following Section~\ref{sec:training-data}, HOI-mix combines SAMP~\citep{hassan2021samp}, CHAIRs~\citep{jiang2023chairs}, and OMOMO~\citep{li2023omomo} motions paired with synthesized RGB videos and curated Bilibili videos with CameraHMR~\citep{patel2025camerahmr} motion estimates. The training set contains 20,812 video--motion--text triplets totaling 29.15 hours, each with 121 frames at 24 fps. These triplets represent 10,847 distinct motion clips, with two appearance variants per interaction motion and one video per Bilibili clip. Motions share a common camera coordinate system; training uses sample-level shuffling across the merged dataset without source-specific reweighting. The legacy validation split contains 30 interaction and 20 non-interaction clips. Interaction data are held out by source sequence, and Bilibili data by clip. The final interaction benchmark contains 100 cases: 12 from CHAIRs, 12 from OMOMO, 6 from SAMP, and 70 additional cases from COUCH~\citep{zhang2022couch}. All benchmark source sequences are disjoint from the current training manifest. However, the benchmark retains the 30 interaction cases from the legacy validation split and is therefore not independent of that split. The separate 35-case ablation set is described in Section~\ref{sec:ablation}.

\noindent\textbf{Implementation Details.} We initialize World2Motion from Cosmos 3 NANO (16B)~\citep{agarwal2026cosmos}, which includes a Qwen3-VL (8B) text backbone~\citep{bai2025qwen3}, 36 Mixture-of-Transformers layers, and a Wan2.2 causal VAE~\citep{wan2025wan}. The motion encoder and decoder are trained from scratch, and the understanding branch remains frozen. We train for 20,000 iterations with AdamW~\citep{loshchilov2019decoupled}, a learning rate of $5\times10^{-5}$, a cosine decay schedule, and 200 warmup steps. Text conditioning is dropped with probability 0.1 for classifier-free guidance~\citep{ho2022classifierfree}. Training uses bfloat16 precision, a batch size of 16, and 16 NVIDIA H100 GPUs, and takes approximately 44 hours.

\noindent\textbf{Baselines.} We compare with text-to-motion methods HY-Motion~\citep{hymotion2025}, Kimodo~\citep{Kimodo2026}, and Go-to-Zero~\citep{fan2025gotozero}; interaction methods LIGHT~\citep{wang2026light} and HOIFHLI~\citep{wu2025hoifhli}; MiniMax-H3~\citep{minimaxh3} + CameraHMR; and CoMoVi~\citep{zhao2026comovi}, fine-tuned on our data. We initialize Kimodo, HOIFHLI, CoMoVi, and World2Motion with poses estimated by CameraHMR~\citep{patel2025camerahmr}. LIGHT and HOIFHLI additionally require 3D object information, while World2Motion uses the scene image.

\subsection{Evaluation}
\label{sec:evaluation}

\noindent\textbf{Motion Quality and Text Alignment.} Following MoMask~\citep{guo2024momask}, we report Fr\'echet Inception Distance (FID) for motion distributions, and R-Precision at rank 1 (R@1) and multimodal distance (MM-Dist) for motion--text alignment. Jerk and foot sliding rate (FSR) follow WHAM~\citep{shin2024wham} and OmniControl~\citep{xie2024omnicontrol}. In Table~\ref{tab:comparison}, jerk uses native frame rates and units of $10\,\mathrm{m/s^3}$; FSR measures the percentage of adjacent frame pairs with sliding on either contacting foot after resampling to 20 fps. Ablation protocols differ (Section~\ref{sec:ablation}); table arrows indicate the preferred direction.

\noindent\textbf{Scene Interaction.} Following OMOMO~\citep{li2023omomo}, we report Contact-F1. We render generated motions in the corresponding 3D scenes and use Qwen3-VL~\citep{bai2025qwen3} to assess whether the requested interaction is completed. Interaction Success Rate (ISR) is the fraction of test cases judged successful. These measures complement motion quality: a smooth motion can match the action description while failing to reach the intended object.

\noindent\textbf{Video--Motion Consistency.} For joint generation methods, we recover motion from the generated video using CameraHMR~\citep{patel2025camerahmr} and compare it with the directly generated motion. PoseError is the pelvis-relative mean Euclidean distance over 22 corresponding body joints, in millimeters. We exclude invalid CameraHMR frames and average within each sequence, then across sequences. This measures cross-modal pose agreement, rather than ground-truth 3D accuracy or agreement of the root trajectories.

\noindent\textbf{Inference Efficiency.} We report mean inference time in seconds. Main-comparison World2Motion outputs use Savitzky--Golay~\citep{savitzky1964smoothing} post-processing (Section~\ref{sec:method}); all ablations disable this filter.

\subsection{Quantitative Comparisons}
\begin{table}[!htbp]
\centering
\small
\caption{Quantitative comparison on 100 interaction cases. FID measures motion distribution quality; R@1 and MM-Dist measure text alignment; Contact-F1 and ISR evaluate scene interaction. Jerk and FSR measure motion stability, and PoseError measures video--motion agreement. Time is mean inference time in seconds. Bold denotes the best reported value per column; dashes indicate unreported scores.}
\label{tab:comparison}
\setlength{\tabcolsep}{2pt}
\fitwidth{\textwidth}{%
\begin{tabular}{l c c c c c c c c c}
\toprule
Method & FID$\downarrow$ & R@1$\uparrow$ & MM-Dist$\downarrow$ & Contact-F1$\uparrow$ & ISR$\uparrow$ & \begin{tabular}[c]{@{}c@{}}Jerk$\downarrow$\\($10\,\mathrm{m/s^3}$)\end{tabular} & \begin{tabular}[c]{@{}c@{}}FSR$\downarrow$\\(\%)\end{tabular} & \begin{tabular}[c]{@{}c@{}}PoseError$\downarrow$\\(mm)\end{tabular} & Time (s)$\downarrow$\\
\midrule
\multicolumn{10}{l}{\textbf{Text-to-motion}}\\
HY-Motion~\citep{hymotion2025} & 0.46 & 0.10 & 1.21 & 0.09 & 0.51 & 3.43 & \textbf{1.25} & -- & 2.83\\
Kimodo~\citep{Kimodo2026} & 0.66 & 0.05 & 1.30 & 0.03 & 0.23 & 35.14 & 6.07 & -- & \textbf{1.93}\\
Go-to-Zero~\citep{fan2025gotozero} & 0.51 & 0.04 & 1.35 & 0.07 & 0.36 & 17.50 & 3.42 & -- & 2.57\\
\midrule
\multicolumn{10}{l}{\textbf{Human--object interaction}}\\
HOIFHLI~\citep{wu2025hoifhli} & 0.82 & 0.02 & 1.37 & 0.09 & 0.12 & 10.62 & 6.28 & -- & 13.94\\
LIGHT~\citep{wang2026light} & 0.62 & 0.03 & 1.35 & 0.17 & 0.46 & \textbf{1.37} & 11.60 & -- & 25.88\\
\midrule
\multicolumn{10}{l}{\textbf{Two-stage video generation and motion recovery}}\\
\begin{tabular}[c]{@{}l@{}}MiniMax-H3~\citep{minimaxh3} +\\CameraHMR~\citep{patel2025camerahmr}\end{tabular} & 0.69 & 0.10 & 1.32 & \textbf{0.24} & \textbf{0.91} & 37.22 & 6.41 & -- & 523.67\\
\midrule
\multicolumn{10}{l}{\textbf{Joint video--motion generation}}\\
CoMoVi~\citep{zhao2026comovi} & 0.71 & 0.12 & 1.25 & 0.09 & 0.69 & 16.10 & 24.45 & 81.28 & 270.45\\
World2Motion & \textbf{0.27} & \textbf{0.24} & \textbf{1.13} & 0.22 & \textbf{0.91} & 8.35 & 2.80 & \textbf{67.50} & 156.59\\
\bottomrule
\end{tabular}
}
\end{table}

\noindent\textbf{Motion Quality and Text Alignment.} In Table~\ref{tab:comparison}, World2Motion achieves the lowest FID and MM-Dist and the highest R@1 among the compared methods. Relative to HY-Motion, FID decreases from 0.46 to 0.27 and R@1 increases from 0.10 to 0.24. It also improves over CoMoVi, whose FID and R@1 are 0.71 and 0.12, respectively.

\noindent\textbf{Scene Interaction.} World2Motion reaches an ISR of 0.91, compared with 0.23--0.51 for the evaluated text-to-motion methods and 0.12--0.46 for the human--object interaction methods. Its Contact-F1 of 0.22 also exceeds those of these baselines. The two-stage pipeline achieves the same ISR and a higher Contact-F1 of 0.24. Successful completion and contact accuracy are therefore distinct: matching the overall interaction success of the two-stage method does not imply identical contact quality. The remaining contact gap motivates more precise supervision for human--object interactions.

\noindent\textbf{Motion Stability.} World2Motion has lower jerk and FSR than Kimodo, Go-to-Zero, HOIFHLI, and the two video-based baselines. HY-Motion achieves lower values on both stability metrics, while LIGHT achieves the lowest jerk but has a higher FSR of 11.60. Thus, the best result depends on which aspect of stability is measured. World2Motion improves stability over the compared video-based pipelines, but it does not surpass every motion-only method. We retain both jerk and FSR to distinguish temporal irregularity from sliding artifacts.

\noindent\textbf{Video--Motion Consistency.} Compared with CoMoVi, World2Motion reduces PoseError from 81.28 to 67.50, a 17.0\% reduction. This indicates closer agreement between its generated video and directly predicted 3D motion under the motion-recovery evaluation. It complements the improvements in text alignment and scene interaction, which assess the generated motion against the prompt and environment rather than against the other output modality.

\noindent\textbf{Inference Efficiency.} World2Motion takes 156.59 seconds on average, compared with 523.67 seconds for MiniMax-H3 + CameraHMR and 270.45 seconds for CoMoVi: speedups of approximately $3.34\times$ and $1.73\times$, respectively. The text-to-motion and human--object interaction methods remain faster. Our efficiency gain concerns video-based motion generation. World2Motion provides both a video and an explicit 3D motion sequence, whereas the faster motion-only baselines do not produce the corresponding video. The results support reducing the cost of obtaining these paired outputs while preserving competitive interaction performance.

\subsection{Qualitative Comparisons}

\begin{figure}[!htbp]
\centering
\includegraphics[width=0.96\linewidth]{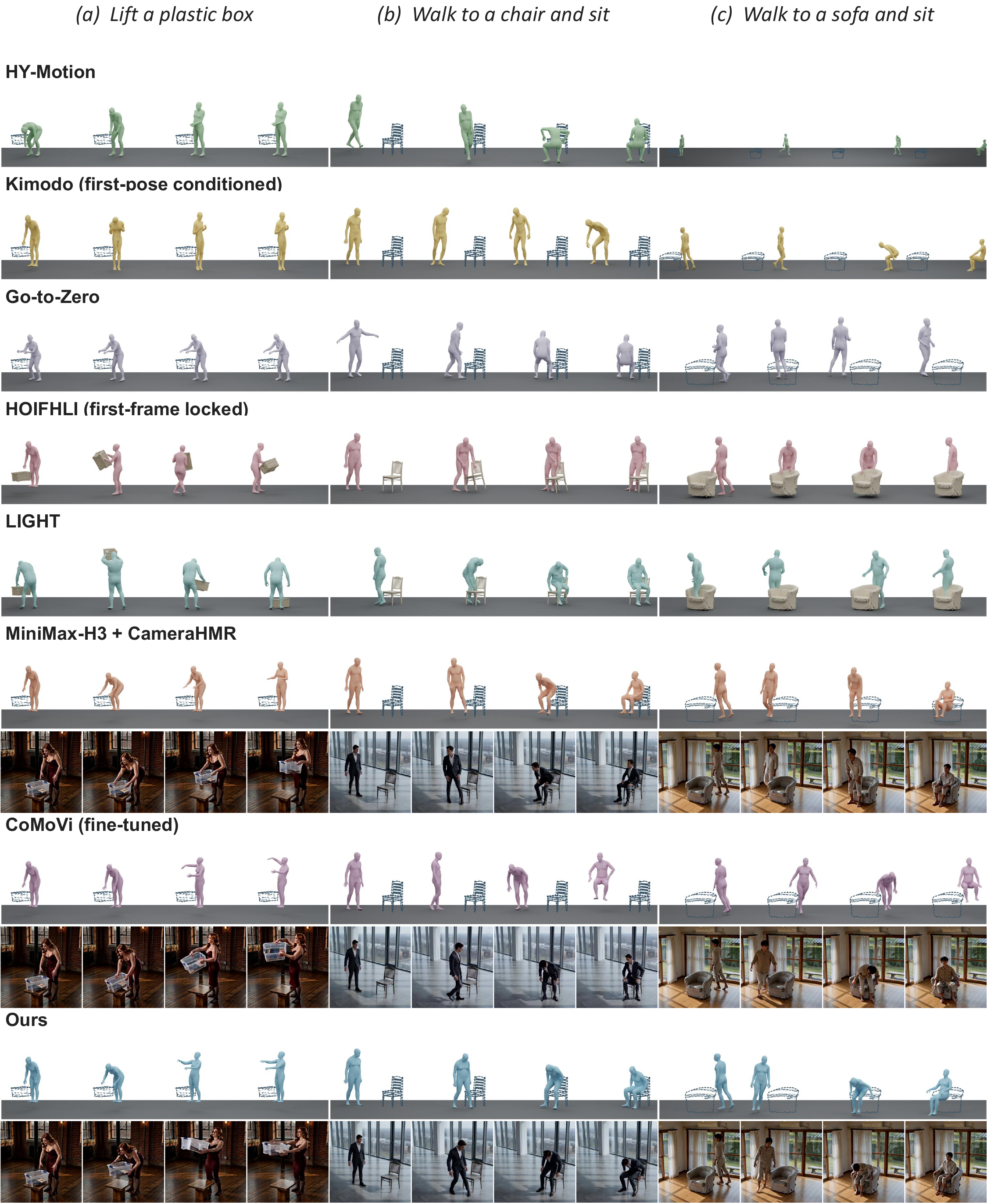}
\caption{Qualitative comparison of scene interactions: (a) lifting a plastic box, (b) walking to a chair and sitting, and (c) walking to a sofa and sitting. Each method shows successive motion states; video-generating methods also show corresponding video frames. Human--object interaction baselines include their predicted object trajectories.}
\label{fig:Comparison}
\end{figure}

Figure~\ref{fig:Comparison} compares actions that depend on both body motion and placement relative to an object. In the sitting examples, producing a plausible sitting pose is insufficient: the character must first approach the intended seat and then align the body with it. The box example also involves coordinating the body with object motion. Viewing the approach and final interaction together helps distinguish action matching from scene-consistent execution. For video-based methods, the paired rows additionally allow the reader to inspect whether the generated video and recovered or directly generated 3D motion describe the same interaction.

\subsection{Ablation Studies}
\label{sec:ablation}

All ablations use 10,000-iteration checkpoints and 35 clips (15 demonstration clips and 20 internet-video clips), without post-processing; Table~\ref{tab:comparison} uses the smoothed 20,000-iteration model on 100 interaction cases. Ablation jerk uses 25 fps. FSR pools sliding observations over contacting foot-joint/frame pairs from four foot joints on a normalized skeleton. Unlike Table~\ref{tab:comparison}, its denominator excludes non-contact observations, so the scores are not directly comparable across protocols. The relative-translation row in Table~\ref{tab:ablation_norm} and the matched $s_m=1$ entry in Table~\ref{tab:ablation_schedule} share the same outputs.

%
\begin{wraptable}{r}{0.45\textwidth}
\centering
\small
\setlength{\belowcaptionskip}{6pt}
\caption{Root translation ablation on 35 cases without post-processing. Raw Jerk uses 25 fps ($10^3\,\mathrm{m/s^3}$). Contact FSR is the sliding fraction among contacting foot-joint/frame pairs on a normalized skeleton (0--1); its definition differs from Table~\ref{tab:comparison}.}
\label{tab:ablation_norm}
\setlength{\tabcolsep}{4pt}
\begin{tabular}{l c c}
\toprule
Translation & \begin{tabular}[c]{@{}c@{}}Raw Jerk $\downarrow$\\($10^3\,\mathrm{m/s^3}$)\end{tabular} & \begin{tabular}[c]{@{}c@{}}Contact FSR $\downarrow$\\(0--1)\end{tabular}\\
\midrule
Absolute & 4.355 & 0.766\\
Relative & \textbf{1.251} & \textbf{0.457}\\
\bottomrule
\end{tabular}
\end{wraptable}
\noindent\textbf{Root translation.} With other settings fixed, relative root translation reduces jerk from 4.355 to 1.251 (71.3\%) and FSR from 0.766 to 0.457 (40.3\%) compared with absolute translation (Table~\ref{tab:ablation_norm}). Both configurations use the same 6D joint-rotation representation. These results support the relative coordinate choice for temporal stability in the tested setting, without root velocity features or post-processing.

\noindent\textbf{Shift-decoupled noise schedule.} In Table~\ref{tab:ablation_schedule}, matched training and inference shifts of $s_m=1$ reduce jerk and FSR by 23.5\% and 28.0\% relative to matched $s_m=10$. Matching shifts gives the lowest values within every row. After training with $s_m=1$, changing only the inference shift to $10$ increases jerk from 1.251 to 8.003. These results support modality-specific shifts and training--inference consistency for the tested stability metrics.

\begin{table}[!htbp]
\centering
\small
\setlength{\belowcaptionskip}{6pt}
\caption{Noise-schedule ablation on the same 35 cases, using Raw Jerk and contact-conditioned FSR as defined in Table~\ref{tab:ablation_norm}. These unfiltered ablation scores are not directly comparable to Table~\ref{tab:comparison}. Columns give the inference motion shift $s_m$; $s_v=10$ is fixed. Bold marks the lowest value per metric within each row.}
\label{tab:ablation_schedule}
\setlength{\tabcolsep}{7pt}
\begin{tabular}{c c c c c c c}
\toprule
 & \multicolumn{3}{c}{Raw Jerk ($10^3\,\mathrm{m/s^3}$) $\downarrow$} & \multicolumn{3}{c}{Contact FSR (0--1) $\downarrow$}\\
\cmidrule(lr){2-4}\cmidrule(lr){5-7}
Training $s_m$ & 1 & 10 & 15 & 1 & 10 & 15\\
\midrule
1 & \textbf{1.251} & 8.003 & 9.703 & \textbf{0.457} & 0.986 & 1.000\\
10 & 7.021 & \textbf{1.635} & 2.557 & 0.985 & \textbf{0.635} & 0.830\\
15 & 8.413 & 2.127 & \textbf{1.809} & 0.981 & 0.700 & \textbf{0.649}\\
\bottomrule
\end{tabular}
\end{table}

\section{Limitations}
\label{sec:limitations}
World2Motion provides a baseline for adapting pretrained video world models to 3D human motion generation, but several limitations remain. First, our training data mainly cover single-person motion and simple interactions, leaving more complex scenarios largely unexplored. Second, despite the improvement in inference speed, World2Motion is still too slow for real-time use. Third, accurate root placement and contact remain challenging. Our evaluation also has limits: a high ISR or low PoseError does not imply precise contact. Future work could expand interaction coverage, incorporate scene geometry and object states, and reduce inference time. For now, generated motions can serve as starting points for further editing. The supplementary video shows examples of agent-assisted smoothing and contact refinement.

\section{Conclusion}
\label{sec:conclusion}
We present World2Motion, which adapts Cosmos 3 to jointly generate scene-aware 3D human motion and video from a single image and a text prompt. We train the model on synthetic video--motion pairs and real videos with estimated motion, and use a shift-decoupled noise schedule to reduce motion jitter. On our interaction benchmark, World2Motion improves motion--text alignment over the compared methods and matches the interaction success rate of the two-stage baseline.

%
%
%
%
%
%
%
%

\FloatBarrier
\bibliography{iclr2027_conference}

@article{Kimodo2026,
  title={Kimodo: Scaling Controllable Human Motion Generation},
  author={Rempe, Davis and Petrovich, Mathis and Yuan, Ye and Zhang, Haotian and Peng, Xue Bin and Jiang, Yifeng and Wang, Tingwu and Iqbal, Umar and Minor, David and de Ruyter, Michael and Li, Jiefeng and Tessler, Chen and Lim, Edy and Jeong, Eugene and Wu, Sam and Hassani, Ehsan and Huang, Michael and Yu, Jin-Bey and Chung, Chaeyeon and Song, Lina and Dionne, Olivier and Kautz, Jan and Yuen, Simon and Fidler, Sanja},
  journal={arXiv:2603.15546},
  year={2026}
}

@article{hymotion2025,
  title={HY-Motion 1.0: Scaling Flow Matching Models for Text-To-Motion Generation},
  author={Tencent Hunyuan 3D Digital Human Team},
  journal={arXiv preprint arXiv:2512.23464},
  year={2025}
}

@inproceedings{tevet2023mdm,
title={MDM: Human Motion Diffusion Model},
author={Guy Tevet and Sigal Raab and Brian Gordon and Yoni Shafir and Daniel Cohen-or and Amit Haim Bermano},
booktitle={The Eleventh International Conference on Learning Representations },
year={2023}
}

@inproceedings{chen2023mld,
  title={Executing your Commands via Motion Diffusion in Latent Space},
  author={Chen, Xin and Jiang, Biao and Liu, Wen and Huang, Zilong and Fu, Bin and Chen, Tao and Yu, Gang},
  booktitle={Proceedings of the IEEE/CVF Conference on Computer Vision and Pattern Recognition},
  pages={18000--18010},
  year={2023}
}

@inproceedings{zhang2023remodiffuse,
  title={Remodiffuse: Retrieval-augmented motion diffusion model},
  author={Zhang, Mingyuan and Guo, Xinying and Pan, Liang and Cai, Zhongang and Hong, Fangzhou and Li, Huirong and Yang, Lei and Liu, Ziwei},
  booktitle={2023 IEEE/CVF International Conference on Computer Vision (ICCV)},
  pages={364--373},
  year={2023},
  organization={IEEE}
}

@article{zhang2024motiondiffuse,
  title={Motiondiffuse: Text-driven human motion generation with diffusion model},
  author={Zhang, Mingyuan and Cai, Zhongang and Pan, Liang and Hong, Fangzhou and Guo, Xinying and Yang, Lei and Liu, Ziwei},
  journal={IEEE transactions on pattern analysis and machine intelligence},
  volume={46},
  number={6},
  pages={4115--4128},
  year={2024},
  publisher={IEEE}
}

@article{jiang2023motiongpt,
  title={Motiongpt: Human motion as a foreign language},
  author={Jiang, Biao and Chen, Xin and Liu, Wen and Yu, Jingyi and Yu, Gang and Chen, Tao},
  journal={Advances in Neural Information Processing Systems},
  volume={36},
  pages={20067--20079},
  year={2023}
}

@inproceedings{zhang2023t2mgpt,
    title={T2M-GPT: Generating Human Motion from Textual Descriptions with Discrete Representations},
    author={Zhang, Jianrong and Zhang, Yangsong and Cun, Xiaodong and Huang, Shaoli and Zhang, Yong and Zhao, Hongwei and Lu, Hongtao and Shen, Xi},
    booktitle={Proceedings of the IEEE/CVF Conference on Computer Vision and Pattern Recognition (CVPR)},
    year={2023},
}

@inproceedings{guo2024momask,
  title={Momask: Generative masked modeling of 3d human motions},
  author={Guo, Chuan and Mu, Yuxuan and Javed, Muhammad Gohar and Wang, Sen and Cheng, Li},
  booktitle={2024 IEEE/CVF Conference on Computer Vision and Pattern Recognition (CVPR)},
  pages={1900--1910},
  year={2024},
  organization={IEEE}
}

@inproceedings{pinyoanuntapong2024bamm,
  title={Bamm: Bidirectional autoregressive motion model},
  author={Pinyoanuntapong, Ekkasit and Saleem, Muhammad Usama and Wang, Pu and Lee, Minwoo and Das, Srijan and Chen, Chen},
  booktitle={European Conference on Computer Vision},
  pages={172--190},
  year={2024},
  organization={Springer}
}

@inproceedings{pinyoanuntapong2025maskcontrol,
  title={Maskcontrol: Spatio-temporal control for masked motion synthesis},
  author={Pinyoanuntapong, Ekkasit and Saleem, Muhammad and Karunratanakul, Korrawe and Wang, Pu and Xue, Hongfei and Chen, Chen and Guo, Chuan and Cao, Junli and Ren, Jian and Tulyakov, Sergey},
  booktitle={Proceedings of the IEEE/CVF International Conference on Computer Vision},
  pages={9955--9965},
  year={2025}
}

@article{SMPL:2015,
  author = {Loper, Matthew and Mahmood, Naureen and Romero, Javier and Pons-Moll, Gerard and Black, Michael J.},
  title = {{SMPL}: A Skinned Multi-Person Linear Model},
  journal = {ACM Trans. Graphics (Proc. SIGGRAPH Asia)},
  month = oct,
  number = {6},
  pages = {248:1--248:16},
  publisher = {ACM},
  volume = {34},
  year = {2015}
}

@inproceedings{kanazawa2018hmr,
  title={End-to-end recovery of human shape and pose},
  author={Kanazawa, Angjoo and Black, Michael J and Jacobs, David W and Malik, Jitendra},
  booktitle={2018 IEEE/CVF Conference on Computer Vision and Pattern Recognition},
  pages={7122--7131},
  year={2018},
  organization={IEEE}
}

@inproceedings{kolotouros2019spin,
  title={Learning to reconstruct 3D human pose and shape via model-fitting in the loop},
  author={Kolotouros, Nikos and Pavlakos, Georgios and Black, Michael J and Daniilidis, Kostas},
  booktitle={Proceedings of the IEEE/CVF international conference on computer vision},
  pages={2252--2261},
  year={2019}
}

@inproceedings{li2022cliff,
  title={Cliff: Carrying location information in full frames into human pose and shape estimation},
  author={Li, Zhihao and Liu, Jianzhuang and Zhang, Zhensong and Xu, Songcen and Yan, Youliang},
  booktitle={European Conference on Computer Vision},
  pages={590--606},
  year={2022},
  organization={Springer}
}

@inproceedings{li2021hybrik,
  title={Hybrik: A hybrid analytical-neural inverse kinematics solution for 3d human pose and shape estimation},
  author={Li, Jiefeng and Xu, Chao and Chen, Zhicun and Bian, Siyuan and Yang, Lixin and Lu, Cewu},
  booktitle={2021 IEEE/CVF Conference on Computer Vision and Pattern Recognition (CVPR)},
  pages={3382--3392},
  year={2021},
  organization={IEEE}
}

@inproceedings{goel2023humansin4d,
  title={Humans in 4d: Reconstructing and tracking humans with transformers},
  author={Goel, Shubham and Pavlakos, Georgios and Rajasegaran, Jathushan and Kanazawa, Angjoo and Malik, Jitendra},
  booktitle={2023 IEEE/CVF International Conference on Computer Vision (ICCV)},
  pages={14737--14748},
  year={2023},
  organization={IEEE}
}

@inproceedings{baradel2024multihmr,
  title={Multi-hmr: Multi-person whole-body human mesh recovery in a single shot},
  author={Baradel, Fabien and Armando, Matthieu and Galaaoui, Salma and Br{\'e}gier, Romain and Weinzaepfel, Philippe and Rogez, Gr{\'e}gory and Lucas, Thomas},
  booktitle={European Conference on Computer Vision},
  pages={202--218},
  year={2024},
  organization={Springer}
}

@article{cai2023smpler,
  title={Smpler-x: Scaling up expressive human pose and shape estimation},
  author={Cai, Zhongang and Yin, Wanqi and Zeng, Ailing and Wei, Chen and Sun, Qingping and Yanjun, Wang and Pang, Hui En and Mei, Haiyi and Zhang, Mingyuan and Zhang, Lei and others},
  journal={Advances in Neural Information Processing Systems},
  volume={36},
  pages={11454--11468},
  year={2023}
}

@inproceedings{patel2025camerahmr,
  title={Camerahmr: Aligning people with perspective},
  author={Patel, Priyanka and Black, Michael J},
  booktitle={2025 International Conference on 3D Vision (3DV)},
  pages={1562--1571},
  year={2025},
  organization={IEEE}
}

@inproceedings{song2026metrichmsr,
  title={MetricHMSR: Metric Human Mesh and Scene Recovery from Monocular Images},
  author={Song, Chentao and Zhang, He and Yuan, Haolei and Lin, Haozhe and Tao, Jianhua and Zhang, Hongwen and Yu, Tao},
  booktitle={Proceedings of the IEEE/CVF Conference on Computer Vision and Pattern Recognition},
  pages={21132--21142},
  year={2026}
}

@article{zhao2026onlinehmr,
  title={OnlineHMR: Video-based Online World-Grounded Human Mesh Recovery},
  author={Zhao, Yiwen and Zheng, Ce and Wang, Yufu and Yang, Hsueh-Han Daniel and Wen, Liting and Jeni, Laszlo A},
  journal={arXiv preprint arXiv:2603.17355},
  year={2026}
}

@misc{minimaxh3,
  title        = {MiniMax H3: An Open Model Breaking the Boundaries Between Tasks and Modalities},
  author       = {{MiniMax}},
  year         = {2026},
  month        = jul,
  howpublished = {\url{https://www.minimax.io/blog/minimax-h3}}
}

@inproceedings{huang2025movein2d,
  title={Move-in-2d: 2d-conditioned human motion generation},
  author={Huang, Hsin-Ping and Zhou, Yang and Wang, Jui-Hsien and Liu, Difan and Liu, Feng and Yang, Ming-Hsuan and Xu, Zhan},
  booktitle={2025 IEEE/CVF Conference on Computer Vision and Pattern Recognition (CVPR)},
  pages={22766--22775},
  year={2025},
  organization={IEEE}
}

@inproceedings{hassan2021posa,
  title={Populating 3D scenes by learning human-scene interaction},
  author={Hassan, Mohamed and Ghosh, Partha and Tesch, Joachim and Tzionas, Dimitrios and Black, Michael J},
  booktitle={Proceedings of the IEEE/CVF Conference on Computer Vision and Pattern Recognition},
  pages={14708--14718},
  year={2021}
}

@inproceedings{hassan2021samp,
  title={Stochastic scene-aware motion prediction},
  author={Hassan, Mohamed and Ceylan, Duygu and Villegas, Ruben and Saito, Jun and Yang, Jimei and Zhou, Yi and Black, Michael J},
  booktitle={Proceedings of the IEEE/CVF International Conference on Computer Vision},
  pages={11374--11384},
  year={2021}
}

@inproceedings{zhao2022coins,
  title={Compositional human-scene interaction synthesis with semantic control},
  author={Zhao, Kaifeng and Wang, Shaofei and Zhang, Yan and Beeler, Thabo and Tang, Siyu},
  booktitle={European Conference on Computer Vision},
  pages={311--327},
  year={2022},
  organization={Springer}
}

@article{li2023omomo,
  title={Object motion guided human motion synthesis},
  author={Li, Jiaman and Wu, Jiajun and Liu, C Karen},
  journal={ACM Transactions on Graphics (TOG)},
  volume={42},
  number={6},
  pages={1--11},
  year={2023},
  publisher={ACM New York, NY, USA}
}

@inproceedings{huang2023scenediffuser,
  title={Diffusion-Based Generation, Optimization, and Planning in 3D Scenes},
  author={Huang, Siyuan and Wang, Zan and Li, Puhao and Jia, Baoxiong and Liu, Tengyu and Zhu, Yixin and Liang, Wei and Zhu, Song-Chun},
  booktitle={Proceedings of the IEEE/CVF Conference on Computer Vision and Pattern Recognition (CVPR)},
  month={June},
  year={2023},
  pages={16750--16761}
}

@article{zou2026infbagel,
  title={InfBaGel: Human-Object-Scene Interaction Generation with Dynamic Perception and Iterative Refinement},
  author={Zou, Yude and Gong, Junji and Gao, Xing and Li, Zixuan and Chen, Tianxing and Zheng, Guanjie},
  journal={arXiv preprint arXiv:2604.04843},
  year={2026}
}

@article{wang2022humanise,
  title={Humanise: Language-conditioned human motion generation in 3d scenes},
  author={Wang, Zan and Chen, Yixin and Liu, Tengyu and Zhu, Yixin and Liang, Wei and Huang, Siyuan},
  journal={Advances in Neural Information Processing Systems},
  volume={35},
  pages={14959--14971},
  year={2022}
}

@inproceedings{jiang2024trumans,
  title={Scaling up dynamic human-scene interaction modeling},
  author={Jiang, Nan and Zhang, Zhiyuan and Li, Hongjie and Ma, Xiaoxuan and Wang, Zan and Chen, Yixin and Liu, Tengyu and Zhu, Yixin and Huang, Siyuan},
  booktitle={2024 IEEE/CVF Conference on Computer Vision and Pattern Recognition (CVPR)},
  pages={1737--1747},
  year={2024},
  organization={IEEE}
}

@inproceedings{wang2024move,
  title={Move as you say, interact as you can: Language-guided human motion generation with scene affordance},
  author={Wang, Zan and Chen, Yixin and Jia, Baoxiong and Li, Puhao and Zhang, Jinlu and Zhang, Jingze and Liu, Tengyu and Zhu, Yixin and Liang, Wei and Huang, Siyuan},
  booktitle={2024 IEEE/CVF Conference on Computer Vision and Pattern Recognition (CVPR)},
  pages={433--444},
  year={2024},
  organization={IEEE}
}

@article{zhao2026ardy,
  title={Autoregressive Diffusion with Hybrid Representation for Interactive Human Motion Generation},
  author={Zhao, Kaifeng and Petrovich, Mathis and Zhang, Haotian and Wang, Tingwu and Tang, Siyu and Rempe, Davis},
  journal={ACM Transactions on Graphics (TOG)},
  volume={45},
  number={4},
  pages={1--14},
  year={2026},
  publisher={ACM New York, NY, USA}
}

@article{bai2025qwen3,
  title={Qwen3-vl technical report},
  author={Bai, Shuai and Cai, Yuxuan and Chen, Ruizhe and Chen, Keqin and Chen, Xionghui and Cheng, Zesen and Deng, Lianghao and Ding, Wei and Gao, Chang and Ge, Chunjiang and others},
  journal={arXiv preprint arXiv:2511.21631},
  year={2025}
}

@article{wan2025wan,
  title={Wan: Open and advanced large-scale video generative models},
  author={Wan, Team and Wang, Ang and Ai, Baole and Wen, Bin and Mao, Chaojie and Xie, Chen-Wei and Chen, Di and Yu, Feiwu and Zhao, Haiming and Yang, Jianxiao and others},
  journal={arXiv preprint arXiv:2503.20314},
  year={2025}
}

@article{agarwal2026cosmos,
  title={Cosmos 3: Omnimodal world models for physical ai},
  author={Agarwal, Niket and Ali, Arslan and Allen, Jon and Antolini, Martin and Aubame, Adeline and Azzolini, Alisson and Bai, Junjie and Bala, Maciej and Balaji, Yogesh and Bapst, Josh and others},
  journal={arXiv preprint arXiv:2606.02800},
  year={2026}
}

@article{zhao2026comovi,
  title={CoMoVi: Co-Generation of 3D Human Motions and Realistic Videos},
  author={Zhao, Chengfeng and Shu, Jiazhi and Zhao, Yubo and Huang, Tianyu and Lu, Jiahao and Gu, Zekai and Ren, Chengwei and Dou, Zhiyang and Shuai, Qing and Liu, Yuan},
  journal={arXiv preprint arXiv:2601.10632},
  year={2026}
}

@inproceedings{jiang2023chairs,
  title={Full-body articulated human-object interaction},
  author={Jiang, Nan and Liu, Tengyu and Cao, Zhexuan and Cui, Jieming and Zhang, Zhiyuan and Chen, Yixin and Wang, He and Zhu, Yixin and Huang, Siyuan},
  booktitle={2023 IEEE/CVF International Conference on Computer Vision (ICCV)},
  pages={9331--9342},
  year={2023},
  organization={IEEE}
}

@article{comanici2025gemini,
  title={Gemini 2.5: Pushing the frontier with advanced reasoning, multimodality, long context, and next generation agentic capabilities},
  author={Comanici, Gheorghe and Bieber, Eric and Schaekermann, Mike and Pasupat, Ice and Sachdeva, Noveen and Dhillon, Inderjit and Blistein, Marcel and Ram, Ori and Zhang, Dan and Rosen, Evan and others},
  journal={arXiv preprint arXiv:2507.06261},
  year={2025}
}

@inproceedings{li2024chois,
  title={Controllable human-object interaction synthesis},
  author={Li, Jiaman and Clegg, Alexander and Mottaghi, Roozbeh and Wu, Jiajun and Puig, Xavier and Liu, C Karen},
  booktitle={European Conference on Computer Vision},
  pages={54--72},
  year={2024},
  organization={Springer}
}

@inproceedings{jung2026dechoi,
  title={Decoupled Generative Modeling for Human-Object Interaction Synthesis},
  author={Jung, Hwanhee and Lee, Seunggwan and Yoon, Jeongyoon and Kim, SeungHyeon and Nam, Giljoo and Huang, Qixing and Kim, Sangpil},
  booktitle={Proceedings of the IEEE/CVF Conference on Computer Vision and Pattern Recognition},
  pages={2253--2263},
  year={2026}
}

@inproceedings{kulkarni2024nifty,
  title={Nifty: Neural object interaction fields for guided human motion synthesis},
  author={Kulkarni, Nilesh and Rempe, Davis and Genova, Kyle and Kundu, Abhijit and Johnson, Justin and Fouhey, David and Guibas, Leonidas},
  booktitle={Proceedings of the IEEE/CVF Conference on Computer Vision and Pattern Recognition},
  pages={947--957},
  year={2024}
}

@article{zhang2026giraf,
  title={GIRAF: Towards Generalizable Human Interactions with Articulated Objects},
  author={Zhang, Xiaohan and Starke, Sebastian and Winkler, Alexander and Bogo, Federica and Aroudj, Samir and Ye, Yuting},
  journal={arXiv preprint arXiv:2607.07880},
  year={2026}
}

@inproceedings{shin2024wham,
  title={Wham: Reconstructing world-grounded humans with accurate 3d motion},
  author={Shin, Soyong and Kim, Juyong and Halilaj, Eni and Black, Michael J},
  booktitle={2024 IEEE/CVF Conference on Computer Vision and Pattern Recognition (CVPR)},
  pages={2070--2080},
  year={2024},
  organization={IEEE}
}

@inproceedings{xie2024omnicontrol,
  title={Omnicontrol: Control any joint at any time for human motion generation},
  author={Xie, Yiming and Jampani, Varun and Zhong, Lei and Sun, Deqing and Jiang, Huaizu},
  booktitle={International Conference on Learning Representations},
  volume={2024},
  pages={28176--28194},
  year={2024}
}

@inproceedings{fan2025gotozero,
  title={Go to zero: Towards zero-shot motion generation with million-scale data},
  author={Fan, Ke and Lu, Shunlin and Dai, Minyue and Yu, Runyi and Xiao, Lixing and Dou, Zhiyang and Dong, Junting and Ma, Lizhuang and Wang, Jingbo},
  booktitle={2025 IEEE/CVF International Conference on Computer Vision (ICCV)},
  pages={13336--13348},
  year={2025},
  organization={IEEE}
}

@inproceedings{wang2026light,
  title={Unleashing guidance without classifiers for human-object interaction animation},
  author={Wang, Ziyin and Xu, Sirui and Zhou, Bing and Gong, Jiangshan and Wang, Jian and Wang, Yu-Xiong and Gui, Liang-Yan and others},
  booktitle={International Conference on Learning Representations},
  volume={2026},
  pages={46121--46141},
  year={2026}
}

@inproceedings{wu2025hoifhli,
  title={Human-object interaction from human-level instructions},
  author={Wu, Zhen and Li, Jiaman and Xu, Pei and Liu, C Karen},
  booktitle={2025 IEEE/CVF International Conference on Computer Vision (ICCV)},
  pages={11176--11186},
  year={2025},
  organization={IEEE}
}

@inproceedings{zhang2022couch,
  title={Couch: Towards controllable human-chair interactions},
  author={Zhang, Xiaohan and Bhatnagar, Bharat Lal and Starke, Sebastian and Guzov, Vladimir and Pons-Moll, Gerard},
  booktitle={European Conference on Computer Vision},
  pages={518--535},
  year={2022},
  organization={Springer}
}

@article{savitzky1964smoothing,
  title={Smoothing and differentiation of data by simplified least squares procedures.},
  author={Savitzky, Abraham and Golay, Marcel JE},
  journal={Analytical chemistry},
  volume={36},
  number={8},
  pages={1627--1639},
  year={1964},
  publisher={ACS Publications}
}

@inproceedings{loshchilov2019decoupled,
  title={Decoupled Weight Decay Regularization},
  author={Loshchilov, Ilya and Hutter, Frank},
  booktitle={International Conference on Learning Representations},
  year={2019},
  url={https://openreview.net/forum?id=Bkg6RiCqY7}
}

@article{ho2022classifierfree,
  title={Classifier-Free Diffusion Guidance},
  author={Ho, Jonathan and Salimans, Tim},
  journal={arXiv preprint arXiv:2207.12598},
  year={2022},
  url={https://arxiv.org/abs/2207.12598}
}
\bibliographystyle{iclr2027_conference}

\clearpage
\appendix
\section{Additional Qualitative Results}

\begin{figure}[!htbp]
    \centering
    \includegraphics[width=0.90\linewidth]{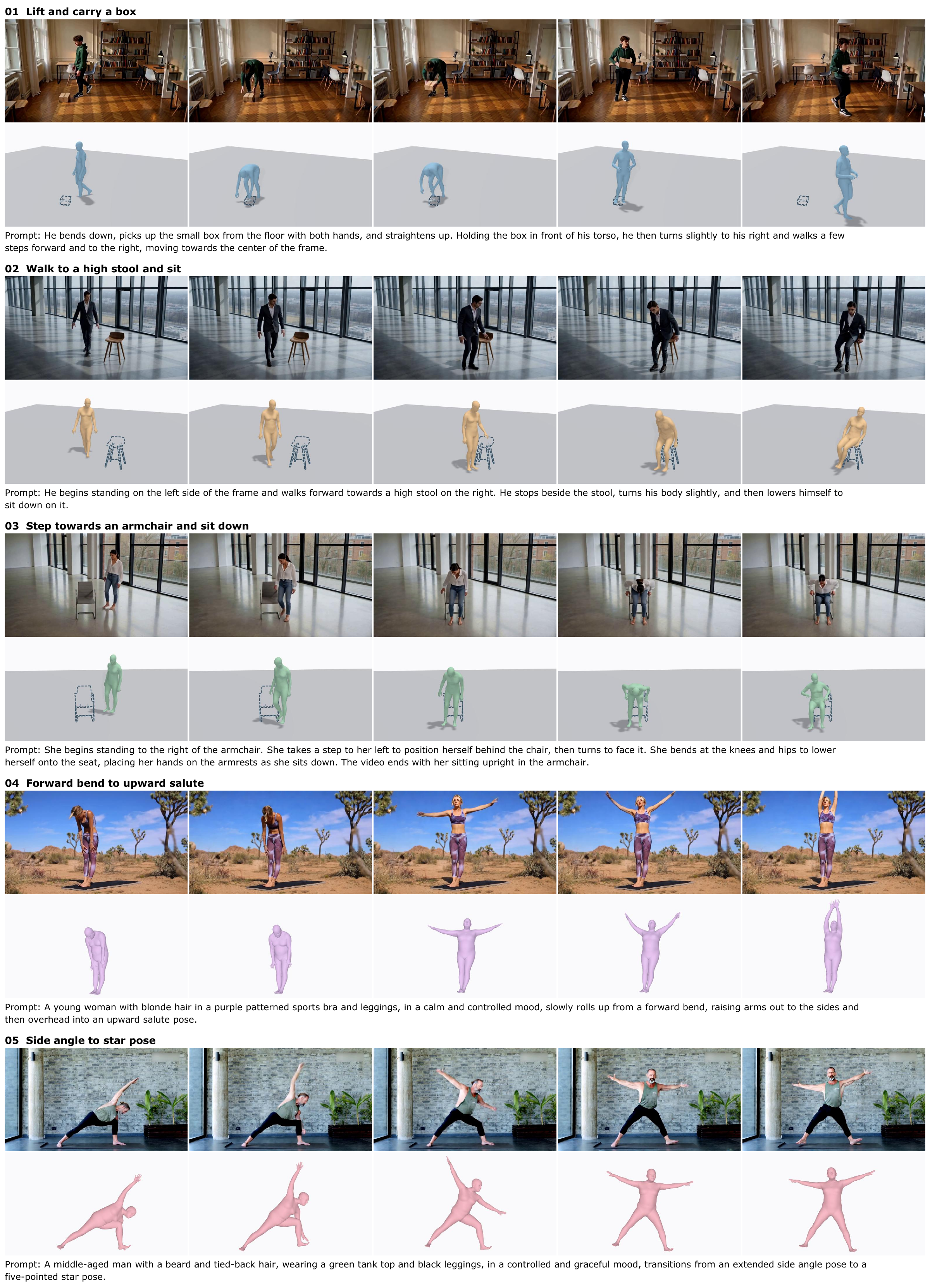}
    \caption{Qualitative results of World2Motion for box lifting and carrying, sitting on a stool or armchair, and two yoga movements. Each example shows five successive video frames (top) and corresponding 3D motion renderings (bottom), ordered from left to right. The input text prompt appears below each example.}
    \label{fig:supplementary-results-1}
\end{figure}

Figure~\ref{fig:supplementary-results-1} presents box lifting and carrying, approaching and sitting on a stool or armchair, and two yoga movements. The sitting examples show the approach to a seat as well as the transition from standing to sitting, allowing the reader to inspect body placement throughout the sequence. The yoga examples show full-body pose changes without object manipulation.


\begin{figure}[!htbp]
    \centering
    \includegraphics[width=0.90\linewidth]{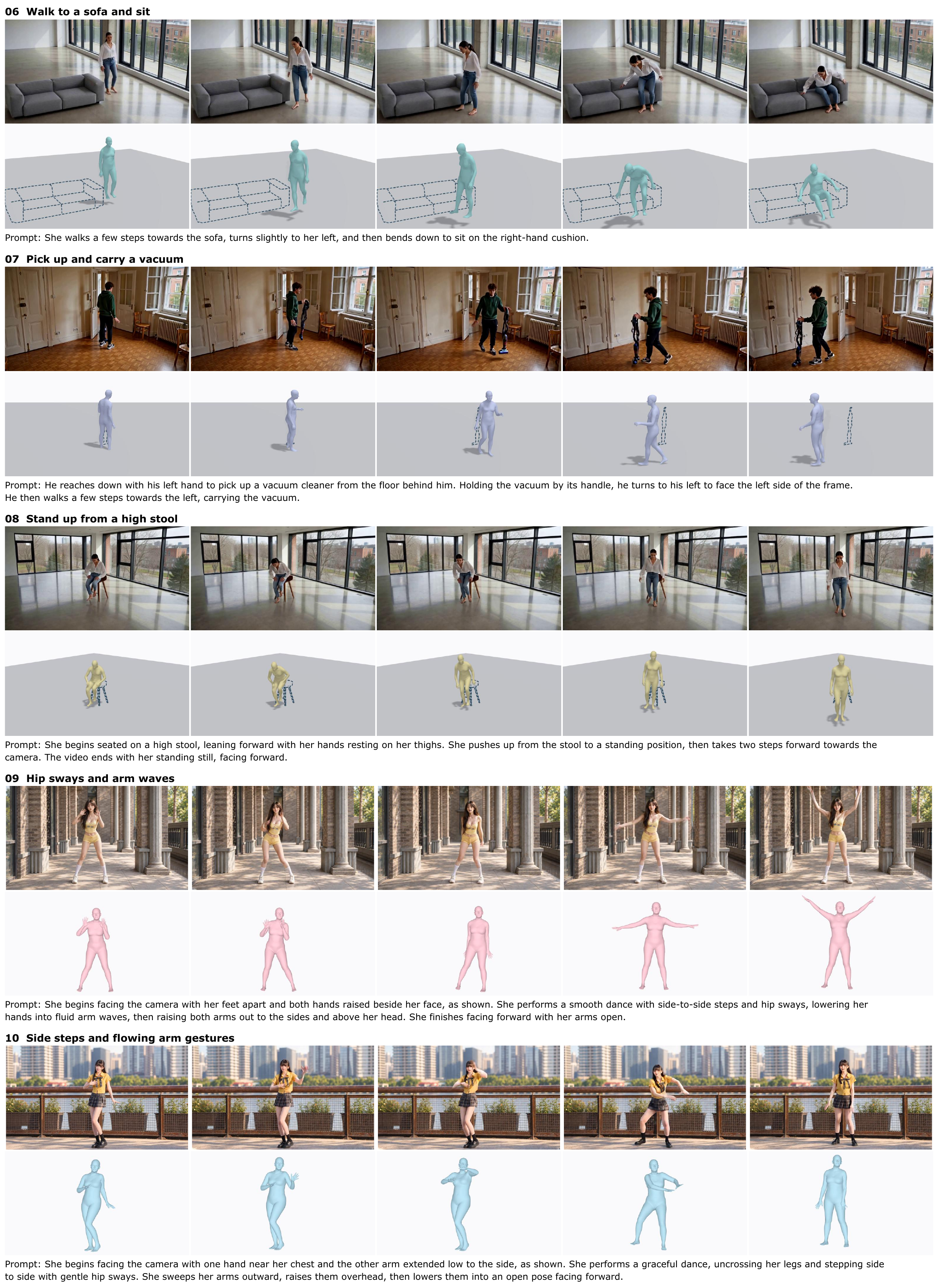}
    \caption{Additional qualitative results of World2Motion for sofa sitting, picking up and carrying a vacuum cleaner, standing up from a stool, and two dance sequences. Each example shows five successive video frames (top) and corresponding 3D motion renderings (bottom), ordered from left to right. The input text prompt appears below each example.}
    \label{fig:supplementary-results-2}
\end{figure}

\clearpage
Figure~\ref{fig:supplementary-results-2} adds sofa sitting, picking up and carrying a vacuum cleaner, standing up from a stool, and two dance sequences. These examples include both changes in body position relative to furniture and coordinated limb movements. In both figures, the paired video and motion rows show five corresponding time points, making it possible to compare how the requested action progresses in the two outputs. The text prompt is provided below each example.

\end{document}